# HMB-GAN: Hybrid Multi-Bézier GAN for Vector Shape Synthesis

Elian Hugh Thiele-Evans, Binh Duong Pham, Hani Omar M Alharbi, Liibaan Aaden, Syed Umer Hasnain Zaidi, and Prem Prakash Jayaraman
*School of Science, Computing and Emerging Technologies*
*Swinburne University of Technology*
Melbourne, Australia
elianhte@gmail.com

Muhammad Saeed, and Boris Eisenbart
*School of Engineering, and School of Design and Architecture*
*Swinburne University of Technology*
Melbourne, Australia

***Abstract*—We explore the use of hybrid quantum-classical generative adversarial networks for synthesising CAD-ready vector geometries. Unlike prior work that operates in rasterised or single-Bézier domains, we introduce HMB-GAN (Hybrid Multi-Bézier GAN), an end-to-end differentiable generative framework that constructs closed shapes through stitched multi-segment Bézier representations with geometric continuity enforced by construction. We compare a quantum-enhanced generator with a classical generator within this architecture and evaluate them across point cloud distribution metrics and geometric shape statistics. Results show that despite faster convergence, a reduction in model parameter count, and slightly improved performance on point cloud metrics, the quantum generator suffers from excessive simulator overhead and thus classically-simulated evaluation suffers from hardware constraints. These results demonstrate the feasibility of modelling structured geometries through hybrid quantum architectures whilst highlighting contemporary hardware limitations.**

***Index Terms*—Quantum Machine Learning, Computer-Assisted Design, Generative Adversarial Networks, Point Cloud Generation, Parametric Bézier Representations**

## 1 Introduction

In modern computer-assisted design (CAD), parametric curves are often used in the construction of objects such as airfoils, part boundaries, and hardware components. Indeed, these curves provide precise, editable geometries which can be optimised for specific structural properties such as aerodynamics, stress tolerance, and flexion.

Relatedly, GANs have demonstrated strong potential for shape generation, reconstruction, and manipulation. Research into GANs for point clouds and 3D meshes has shown that adversarial learning can capture spatial dependencies without relying on explicit parametric surfaces [1]. Similarly, in 2D and CAD design, models based on Bézier curves have been used to represent complex forms compactly while maintaining mathematical smoothness and structural coherence [2]. Integrating Bézier-based geometric representations within GAN frameworks allows the generator to produce continuous, editable design outputs. This is an advantage over pixel-based synthesis for engineering and design applications.

More recently, the emergence of quantum and hybrid quantum-classical GANs has introduced new opportunities for generative modelling. Quantum GANs (QGANs) utilise variational quantum circuits as generators, leveraging quantum superposition and entanglement to represent probability distributions that may be intractable for classical networks [3]. Hybrid architectures combine quantum generators with classical discriminators to improve expressivity in modelling while keeping the whole system computationally practical. These approaches have shown early promise in tasks involving limited or complex datasets, where quantum embeddings may enhance sample diversity and generalisation [4].

Nevertheless, challenges remain in applying GANs to geometric domains. Standard GAN architectures are primarily designed for grid-based data such as images, whereas Bézier curves and 3D point clouds do not fit neatly into a grid. While QGANs offer theoretical advantages, their practical integration into hybrid systems for design geometry is still underexplored.

This research therefore seeks to address this gap by developing a hybrid quantum-classical generative adversarial framework for synthesising geometric data using Bézier-based representations. Building upon the foundational GAN architecture introduced by [5] and the recent advancements in quantum variational circuits, this study aims to explore the impact quantum feature encoding may have within adversarial learning on geometric diversity, and stability.

The main contributions of this paper are as follows:

1) Exploration of small-scale quantum generator feasibility on classical hardware, highlighting simulation constraints and GAN convergence implications.
2) An end-to-end differentiable GAN architecture that constructs closed, multi-segment Bézier curves in vector space, without relying on rasterisation or symbolic primitives.
3) A Bézier decoder that creates CAD-ready shapes via geometric stitching constraints, which enforces continuity and closure across segments.

## 2 Related Work

Existing work on geometric modelling and vector graphic synthesis can broadly be categorised into three approaches: rasterisation-based models, sequence-based SVG generation

Published in 2026 IEEE International Conference on Quantum Software (QSW). DOI: 10.1109/QSW72780.2026.00033

models, and parametric curve representations. Differentiable rasterisation methods (e.g. DiffVG [6]) are potent, but they *indirectly* optimise vector graphics, which can make them poorly suited for tasks that require structured and consistent geometries, and do not naturally enforce closure or continuity. Sequence-based models operate directly on SVG parameters (e.g. DeepSVG, SketchRNN [7], [8]), but do not have closure or geometric continuity guarantees, and do not have per-segment-level correspondence across samples. Existing parametric approaches to generating geometries model parameters directly in latent space, but are limited in that they can only represent single-segment shapes (e.g. BézierGAN [9]) and therefore are not directly extensible to a structured multi-segment shape setting.

Thus, there remains a gap in generative modelling frameworks that can produce structurally coherent, complex two-dimensional CAD-ready geometries.

## 3 Methods

### 3.1 Proposed Framework

HMB-GAN is composed of three components: (1) a hybrid quantum generator utilising VQCs, (2) a structured multi-segment Bézier decoder that outputs closed vector shapes, and (3) a PointNet discriminator that operates over point clouds. Figure 1 displays the overall GAN architecture for the framework.

Random latent vectors are passed through the generator, and transformed into higher-dimensional latents, either through non-linear mapping (classical comparison), or through multivariate feature encoding through VQCs. A decoder maps the higher-dimensional latent vectors into structured Bézier segments, which are sampled into point clouds and subsequently classified by the discriminator. Gradients from the discriminator backpropagate through the decoder and generator, encouraging more realistic shapes.

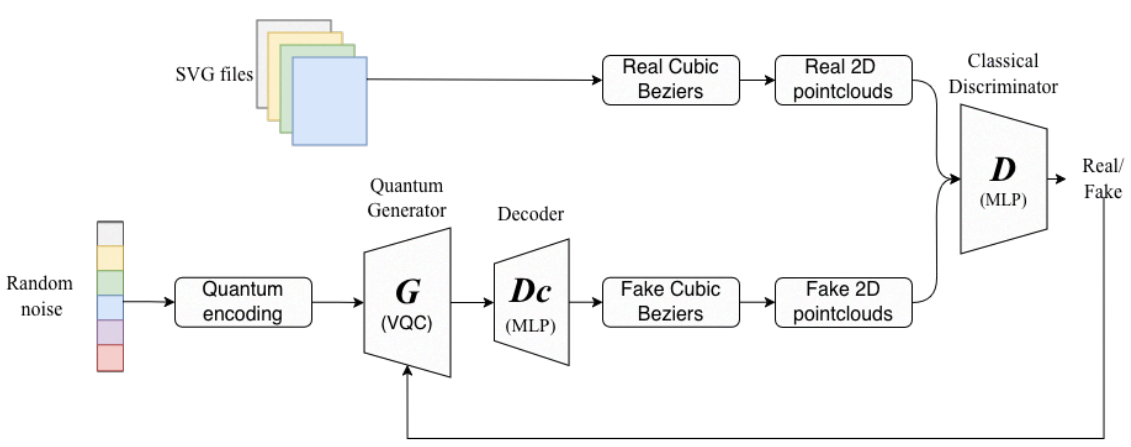

Figure 1. Quantum Training Pipeline

### 3.2 Quantum Generator

The quantum generator encodes latent codes $z$ into quantum states for consumption by variational quantum circuits (VQCs). The noise is transformed into forms that capture the mechanics of the empirical data distribution. To promote smooth transformations and stable gradients, the generator uses scaled-residual skip connections [10]. Given the classic instability of adversarial training, incremental learnable deviations help stabilise the min-max game, whereas by contrast, feature reparameterisations can be volatile.

Quantum $\mathcal{G}$ encodes latent variables into qubits, and applies parameterised RZ, RY, RY rotations to enrich the feature space. A CZ entanglement gate is applied to allow neighbouring qubits to capture non-linear dependencies. The final result is converted back into a classical form for ingestion by the next layer of the generator.

The quantum $\mathcal{G}$ is defined by the recursive structure and quantum parameterisation given in (1),

$$h_0 = \mathrm{LN}(W_0 z), h_{k+1} = h_k + \alpha Q_{k(h_k;\theta_k)}, k = 0..3 \quad (1)$$

where $\theta_k$ represents the trainable rotation parameters and entanglement gate angles of the VQC, and alpha is a fixed scaling factor for residual magnitude.

Subsequently, parameterised rotations and entangling layers are applied. The current circuit uses data-reuploading [11] to re-encode $x$ before each variational layer, capturing richer relationships between latent dimensions with fewer qubits, and improving gradient flow. The circuit parameter $\theta$ is initialised near zero, to stabilise early training of the GAN.

The final output is computed as (2)

$$\mathcal{G}(z) = W_{\text{out}} h_4 \quad (2)$$

Thus, features from $\mathcal{G}$ are passed directly to the decoder to extract latent Bézier representations.

### 3.3 Decoder

Dc is formulated as a shallow network with two heads, one that expands generator features into Bézier space, and another that uses generator features to control the curvature of the Bézier handles.

To achieve geometric closure, Dc stitches the segments so that the final $P^3$ point is equal to the first segment's $P^0$ point in Cartesian space. Moreover, to ensure $C^0$ continuity, segments are stitched to ensure each segment's $P^3$ point is shared with the subsequent segment's $P^0$. In practice, this means that the resulting tensor indices are strided. As with previous research into shape generation [8], we have found that using a mean-centred cumulative sum of control points imparts a strong geometric continuity prior. Using control point deltas (as opposed to absolute coordinates) allows the network to focus on Bézier transformations rather than detangling the segments at initialisation.

Dc is composed of an MLP with a fully connected layer, layer normalisation, SiLU activation and a final fully connected layer that outputs Bézier segments. To handle learned curvature, a second MLP includes two fully connected layers with a SiLU activation, with weights initialised to produce smooth curves at the onset. The positional logits of the curvature head are passed through sigmoid activations (denoting how far along the chord the control point is), and the curvature deviation logit is passed through a tanh activation (how strongly the Bézier handle deviates from the chord). Cyclical encodings of segments are used as positional encodings that impart a smooth connectivity prior between segments. Finally, the stitched tensor is per-sample bounding box normalised $[-1, +1]$ to match downstream shape expectations.

We define a vector decoder in (3) as

$$\mathrm{Dc} : \mathbb{R}^{d_h} \to \mathbb{R}^{S*4*2}, \quad (3)$$

with stitching enforced as defined in (4):

$$c_i = P_0^{i+1} - P_0^i, P_3^i = P_0^i + c_i, P_3^S = P_0^1 \tag{4}$$

and curvature parameterised as in (5):

$$C_k^i = P_0^i + \sigma(s_{k,i})c_i + \tanh(\delta_{k,i})n_i, k \in \{1, 2\}, \tag{5}$$

where segment position $s$ is encoded as in (6)

$$e_s = \left[\cos\left(2\pi \frac{s}{n} k\right), \sin\left(2\pi \frac{s}{n} k\right)\right]_{k=1}^{K} \tag{6}$$

where $n$ is the total number of Bézier segments and $K$ is the number of harmonic components.

Unlike single-curve parametric models [9], our model represents each shape as a fixed set of $N > 1$ stitched Bézier segments. This preserves per-segment correspondence across samples, whilst combining cumulative stroke generation [8] with learned Bézier handle parameterisations [9].

Through gradients provided by the discriminator, Dc helps guide the Bézier segments composed by $\mathcal{G}$.

### 3.4 Discriminator

The discriminator is formulated in the DeepSets/PointNet [12], [13] style, wherein set data features are extracted, before pooling through a cardinality invariant operation (e.g. mean, max, or log-sum-exp), and finally features are further extracted to emphasise set-to-set differences. During training, the order of points within each set is randomly permuted to prevent accidental reliance on set ordering.

To promote scale invariance and translation invariance, we introduce a set normalisation module ("SetNorm"). The intended outcome is to allow the discriminator to focus on geometric cues rather than the more superficial scale and translation cues that carry louder signals. Concretely, SetNorm centres and rescales point clouds to unit RMS magnitude, followed by a learned affine bias and gain (analogous to LayerNorm or BatchNorm). As signal is lost from the removal of these cues, the affine transformation acts as a learned magnitude amplification. Formally, the SetNorm operation normalises each point cloud as defined in (7)

$$\tilde{x} = x - \overline{x}, y = \frac{\tilde{x}}{\sqrt{\overline{\tilde{x}^2} + \varepsilon}} \gamma + \beta \tag{7}$$

where $\tilde{x}$ is the point cloud centroid and $\overline{\tilde{x}^2}$ denotes the mean squared magnitude across points and coordinates.

The discriminator consists of the SetNorm module, three spectrally normalised fully-connected layers and leaky ReLU activations. As per [14], mix pooling is used via concatenation of mean and max values of the feature tensor. Finally, following layer normalisation, two fully-connected layers and a leaky ReLU activation downsample to a single logit representing real and fake probabilities.

Unlike PointNet, and other DeepSet style discriminators, we apply spectral normalisation to all the linear layers in the discriminator to stabilise training via a 1-Lipschitz constraint under the SN-GAN formulation [15]. Gradients from D are backpropagated through the vector decoder and generator, promoting more realistic shapes.

### 3.5 Loss Architecture

Our goal is to learn a mapping of latent representation $z$ to output $x$ that captures both the statistical distribution of real samples, as well as their geometric fidelity. To this end, our network incorporates three types of losses:

1) adversarial losses, operationalised as adaptive-margin hinge loss [16]
2) auxiliary losses, including feature matching [17] and mode-seeking loss [18]
3) geometric regularisation losses, consisting of Chamfer distance [19], quantile-based shoelace-area penalty [20], [21], and $G^1$ continuity penalty loss

The discriminator is trained with the corresponding adaptive-margin hinge objective only.

$G^1$ continuity promotes tangent direction matching across segment joins. This discourages oscillatory segments, as the network is rewarded for producing smooth curvatures. We define the $G^1$ loss term as (8):

$$\ell_{G^1} = \frac{1}{J} \sum_{j=1}^{J} \left(1 - \frac{v_1^{(j)} \cdot v_2^{(j)}}{\|v_1^{(j)}\| \|v_2^{(j)}\|}\right) \tag{8}$$

The shoelace area loss (9), derived from Greene's theorem [20], [21], penalises synthetic Béziers whose unsigned area falls below the 10th percentile of empirical area sizes. This discourages degenerate shapes such as hairpin chords or back-looped geometries. Though the 10th percentile cutoff is a heuristic, it was empirically chosen so as to not punish shapes that are feasibly small. The shoelace loss term is expressed as the following:

$$\ell_{\text{area}} = \mathbb{E}_{\text{fake}}\left[\max\left(0, \tau_q - \tfrac{1}{2} \left| \oint_c (x \text{ dy} - y \text{ dx}) \right| \right)\right], \tag{9}$$

where $\tau_q$ denotes the quantile of the unsigned areas of real samples.

The full objective for $\mathcal{G}$ is defined through three loss components: auxiliary loss (10), geometric loss (11), and the final generator objective (12), where the auxiliary and geometric losses are weighted accordingly:

$$\ell_{\text{aux}} = \lambda_{\text{FM}} \ell_{\text{FM}} + \lambda_{\text{MS}} \ell_{\text{MS}} \tag{10}$$

$$\ell_{\text{geo}} = \lambda_{\text{CH}} \ell_{\text{CH}} + \frac{\lambda_{\text{CH}}}{10} \ell_{\text{area}} + \frac{\lambda_{\text{CH}}}{10} \ell_{G^1} \tag{11}$$

$$\ell_{\text{full}} = \ell_{\mathcal{G}} + \ell_{\text{aux}} + \ell_{\text{geo}} \tag{12}$$

Here, $\ell_{\mathcal{G}}$ denotes the generator adversarial term, $\ell_{\text{FM}}$ and $\ell_{\text{MS}}$ denote feature-matching and mode-seeking losses, and $\ell_{\text{CH}}$, $\ell_{\text{area}}$ and $\ell_{G^1}$ denote the Chamfer, shoelace-area, and tangent-continuity terms respectively.

### 3.6 Training

Training follows the SN-GAN approach with AMP and EMA stabilisation [15]. Hyperparameters were chosen using TPE-based Bayesian optimisation [22] against 1-Nearest Neighbour Accuracy (1-NNA) scores. Final model evaluation is reported across distribution metrics to ensure improvements are not due to over-optimisation of a single metric.

## 4 Evaluation

### 4.1 Dataset

The current dataset is the SketchGraphs dataset [23] collated by Princeton University. It is a collection of 15 million sketches extracted from real-world CAD models coupled with an open-source data processing pipeline.

Sketches are then converted to SVGs via node traversal, with an additional constraint that only single-stroke shapes with fully closed paths are retained. The point clouds are extracted from each SVG and canonicalised through centering, normalisation, and PCA-based rotation to the primary axis, followed by uniform arc-length resampling to 256 points. The arc resampling step ensures an equal distribution of points across any given shape, avoiding density biases where shorter segments may receive disproportionately higher points.

### 4.2 Evaluation Metrics

We evaluate point cloud distribution metrics and shape metrics to provide analysis of geometric quality, mode coverage, and shape validity. We evaluate distributional similarity via Chamfer-based minimum matching distance (MMD), coverage (COV), and 1-Nearest Neighbour Accuracy (1-NNA), as described by PointFlow [24]. Lower MMD scores indicate geometric similarity between sets, higher COV indicates better mode coverage, and 1-NNA scores closer to 50% indicate less separability between real and synthetic samples.

To assess shape validity, we also report enclosed area, $G^1$ angle, self-intersection rate, and closure error.

### 4.3 Baseline and Comparative Setup

To examine our generative framework, we compare the quantum GAN with a classical version wherein the generator residual mappings are implemented using fully connected layers instead of VQCs. All training dynamics are otherwise identical, including GAN architectural structure, number of layers, and so on.

Our architecture introduces a generative framework for multi-Bézier parametric shapes with structured geometric continuity and closure. To our knowledge, few existing generative architectures produce such shapes end-to-end in vector space with built-in geometric constraints. In order to use existing models for comparison, such as point cloud GANs, the training pipeline would need restructuring to be compatible, and would not be a fair comparison. Consequently, suitable architectural baselines are difficult to construct. To contextualise GAN performance, we evaluate via two methods:

(1) A baseline upper-bound on distribution metrics via repeated random splitting and subsequent evaluations of the test set ($N = 10$).
(2) ablations of the GAN architecture which isolate the contributions of adversarial, auxiliary, and geometric losses.

### 4.4 Experimental Setup

In order to estimate the inter-run variability of model results, experiments are conducted across three independent training runs using random seeds, for a total of six runs between the quantum and classical models. As adversarial training is largely stochastic, evaluations across multiple independent runs allow us to estimate the variance in the training itself.

To estimate the inter-data variability across evaluations, we perform bootstrapped evaluations of MMD and COV ($N = 500$) for each training run. As 1-NNA is sampling dependent on the dataset split, bootstrap sampling is not appropriate. Instead, we opt for Monte-Carlo subsampling, with a subsample fraction of 0.9. As a result, 95% intervals for 1-NNA represent sub-sample variability rather than evaluation set variability.

The dataset uses a standard train-validation-test split of 60/20/20. We use a multilabel stratified shuffle split using approximate pseudolabels derived from point-cloud similarity metrics. These labels are used only to encourage a balanced distribution of shape classes, and not for evaluation labels. The validation set is used to monitor training diagnostics. All reported metrics, however, are computed against the test set exclusively.

All training runs occur for 1000 epochs under identical training regimes and hyperparameters, barring the differences in the quantum and classical generator implementations.

## 5 Results and Discussion

### 5.1 Distribution and Shape Validity Analysis

As with prior research in the point cloud domain, the metrics proposed by [24] are used to evaluate the synthetic generator samples. Through these point cloud metrics, we can assess whether the quantum model produces synthetic distributions that are comparable or better (in the context of mode coverage, geometric fidelity, and separability) than the classical baseline. Furthermore, these metrics allow us to appraise whether the framework is successfully modelling the data distribution of the training dataset, and thereby capable of synthesising detailed, diverse, and geometrically realistic two-dimensional shapes. We additionally examine model size and parameter count to compare the computational efficiency of the classical and quantum generators. Table I explores the distribution metrics for the model, computed on a held-out test set.

Both generators were able to achieve favourable metrics, with the quantum generator marginally outperforming the classical variant on all three point cloud measures. Moreover, the quantum generator had a marked reduction in parameter count (1,847,386) in comparison to the classical model (3,944,394). Interestingly, both models, whilst attaining worse coverage than the null model, were able to achieve a lower MMD score than baseline.

In addition to distribution similarity, we also examine the geometric validity of the synthetic shapes. Assessing the geometric validity of the synthetic shapes is important (e.g. self-intersections, geometric closure), as the framework is intended for CAD use cases that depend on structural guarantees. Moreover, we seek to analyse whether the quantum model accurately produces valid shapes when compared to the classical baseline. Table II presents the shape validity metrics for the trained models and training set.

TABLE I
DISTRIBUTION SIMILARITY

| Model | # Parameter | MMD (↓) | COV (%, ↑) | 1-NNA (%, ↓) |
|---|---|---|---|---|
| Classical | 3.94M | $0.064 \pm 0.00$<br>$[0.07 - 0.07]$ | $50.57 \pm 0.40$<br>$[49.05 - 52.59]$ | $65.9 \pm 0.8$<br>$(57.5 - 58.7)$ |
| Quantum | **1.85M** | $\mathbf{0.059} \pm 0.00$<br>$[0.06 - 0.07]$ | $\mathbf{50.96} \pm 1.18$<br>$[50.11 - 53.51]$ | $\mathbf{63.2} \pm 2.4$<br>$(55.54 - 56.7)$ |
| Training Set | - | $0.064 \pm 0.00$ | $\mathbf{56.22} \pm 0.80$ | - |

*All metrics reported as mean ± standard deviation across 3 independent training runs. Best score is bolded. Brackets denote 95% confidence intervals via bootstrapping for MMD and COV. Parentheses denote 95% intervals obtained via Monte-Carlo subsampling for 1-NNA.*

TABLE II
SHAPE VALIDITY METRICS

| Model | Area | $G^1$ Angle (°, ↓) | Self-Intersection (%, ↓) | Closure Error (↓) |
|---|---|---|---|---|
| Classical | $0.940 \pm 0.083$ | $48.58 \pm 1.78$ | $41.51 \pm 2.15$ | $0.00 \pm 0.00$ |
| Quantum | $0.639 \pm 0.029$ | $49.23 \pm 5.75$ | $\mathbf{37.96} \pm 5.28$ | $0.00 \pm 0.00$ |
| Training Set | $1.480 \pm 0.880$ | $72.35 \pm 42.76$ | 8.87 | $0.01 \pm 0.13$ |

*All metrics reported as mean ± standard deviation across 3 independent training runs. Best score is bolded.*

Both GAN variants present zero closure error, as shapes are constructed as closed stitched Bézier segments. The quantum generator exhibited fewer shapes with self-intersecting curves in comparison to the classical generator. Both models achieved comparable smoothness across generated curves. However, the test set exhibits higher variability across both area and smoothness metrics in comparison to the generative models, indicating that the generators tended to produce more regularised geometries.

Improved performance from the quantum variant may not stem purely from inherent quantum expressivity and more so from the capture of multivariate correlations. The flat MLP approach of the classical generator relies on learning correlations as an emergent property, whereas the quantum generator implicitly encodes multivariate relationships. Therefore, the advantage of the quantum approach may be better structured as the advantage of a multivariate inductive bias rather than purely quantum computation.

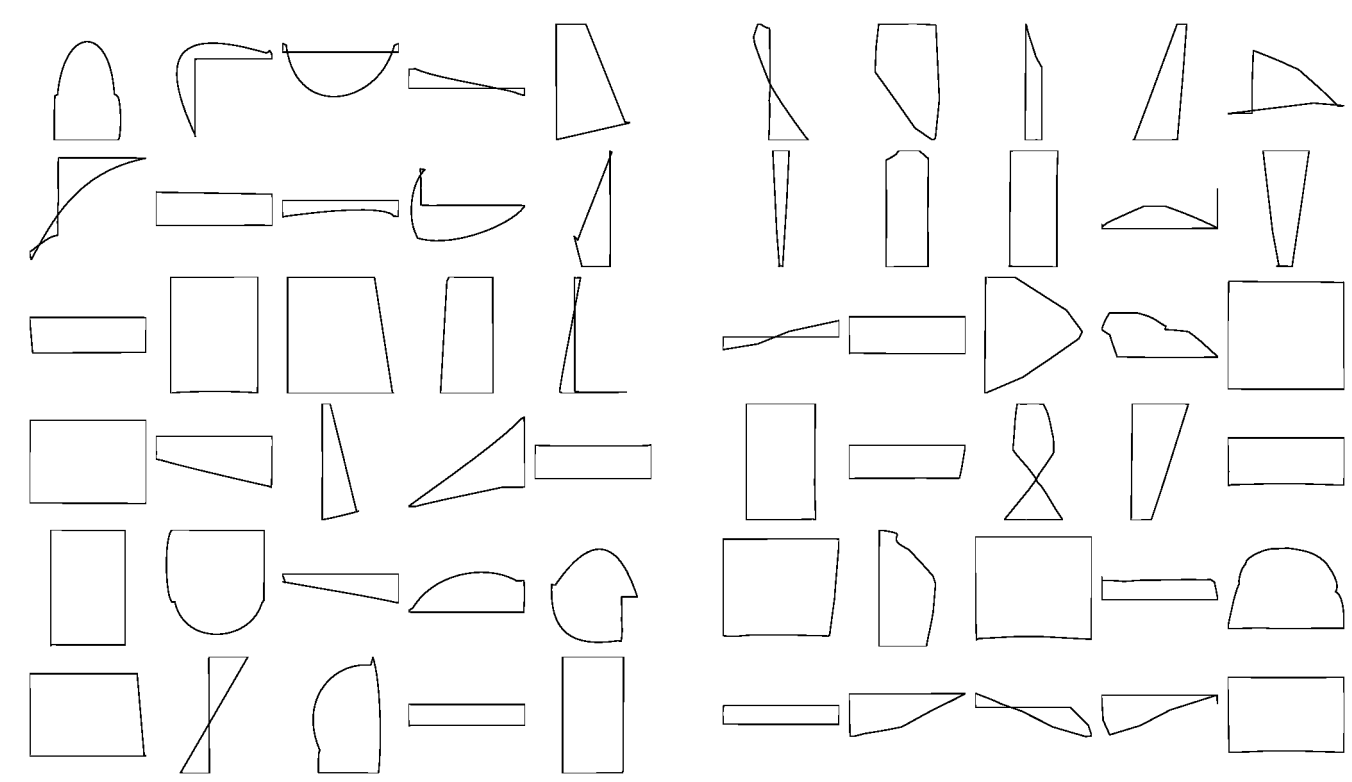

(a) Classical Shape Outputs (b) Quantum Shape Outputs

Figure 2. Classical and Quantum Shape Outputs

The quantum synthetic shapes exhibit earlier complexity than their classical counterparts; the quantum samples provided more visually interesting shapes. This notably occurs at epoch 500 compared to the classical epoch 1000, representing faster convergence on the quantum generator's part.

### *5.2 Failure Modes*

Symmetry appears to be a particularly difficult shape descriptor to emulate, likely because it requires precise control over all Béziers to produce mirrored structures. Overall, this suggests that though local curvature was learned by the network, global structure was difficult to capture. Although we see evidence of global structure learned later in training (e.g. early signs of symmetry), further training risks collapse as well as diminishing returns in the adversarial game.

Indeed, the network exhibits difficulty in capturing the broader distribution of shapes. However, it is inaccurate to state that this purely due to the GAN's lack of representational capacity. As per Table I, the null model's upper-bound on coverage is 56.2%, where by comparison the classical model achieved 50.6%. Thus, limitations in mode capture are equally likely to be the result of dataset limitations.

### *5.3 Training Efficiency and Stability*

On an HPC NVIDIA A100 node, the classical GAN ran for approximately 1.12 minutes per epoch. By comparison, the quantum variant on the GPU was unacceptably slow at 87 minutes per epoch, and 3.54 minutes per epoch on CPU.

The classical GAN ran with moderate RAM usage and low CPU and GPU usage. This is likely due to the GAN architecture relying on many small kernels (such as the geometric reconstructive terms), as well as being limited by smaller batch sizes. Whilst larger batch sizes were feasible for the GPU (and indeed upwards of 1024-sized batches could easily fit in memory), the result was overly smoothed gradients and homogeneous shapes.

The quantum CPU variant used lower RAM and higher CPU usage than the classical run, consistent with the loss of GPU compute. Overall, the compute resources show that the quantum generator runs much faster on CPU with the PennyLane simulator.

This drastic difference in computation times between quantum GPU and CPU is likely due to the PennyLane simulator prioritising parallel computations of observables (e.g. >20 Qubits exhibiting superior GPU performance [25]), over parallel broadcasting over batches [26]. This would also explain the disparity between classical and quantum runs, as the classical generator was able to rely on Torch's batched kernels and vectorised tensor operations.

### 5.4 Cross-Comparison

To analyse the role adversarial training has in the performance of the current model, ablation studies were performed on the classical model (due to the quantum model being computationally expensive). The auxiliary and reconstructive terms were disabled to examine how the model performs without the stabilising terms – see Table III.

TABLE III
CROSS-COMPARATIVE STUDY RESULTS

| Model | MMD (↓) | COV (%, ↑) | 1-NNA (%, ↓) |
|---|---|---|---|
| Full model | $0.064 \pm 0.00$ | $\mathbf{50.57} \pm 0.40$ | $65.91 \pm 0.83$ |
| Auxiliary Ablation | $0.061 \pm 0.00$ | $50.34 \pm 0.90$ | $66.95 \pm 0.83$ |
| Geometric Ablation | $0.064 \pm 0.00$ | $50.19 \pm 0.29$ | $65.96 \pm 0.54$ |
| Adversarial Only | $\mathbf{0.057} \pm 0.00$ | $49.46 \pm 0.33$ | $\mathbf{65.19} \pm 1.40$ |

*All metrics reported as mean ± standard deviation across 3 independent training runs. Best score is bolded.*

Ablations of the various regularising terms reveal the efficacy of the additional auxiliary and geometric terms. Specifically, the removal of all additional loss terms in the adversarial-only case resulted in the best MMD ($0.057 \pm 0.00$) and 1-NNA scores ($65.189\% \pm 1.40\%$). By comparison, removal of the geometric terms resulted in an increase of the coverage metric, which is expected when geometric fidelity is de-prioritised. Additionally, ablation of auxiliary terms resulted in a degradation of coverage and 1-NNA scores, albeit marginally outperforming the geometric ablation in MMD. It was expected that geometric terms act as regularisers of geometric fidelity, which may influence the network to approach more average modes (the penalty against coverage which the auxiliary terms successfully offset). The full model, as would be expected, was middling, with average point cloud metrics across the board.

Despite a decrease in point cloud metrics, the network exhibits an increase in geometric errors when geometric terms are ablated. Indeed, this degradation in performance reflects the perceptual differences of the synthetic shapes, wherein the geometries tended to exhibit more local artefacts. The full model, therefore, provides a more balanced objective than the ablative models, sacrificing some distributional fidelity for improved geometry.

### 5.5 Limitations

It should be noted that the current quantum generator is evaluated under classical simulation. Thus, the runtime characteristics represent simulator overhead rather than physical quantum hardware, and the results should be interpreted as a proof-of-concept exploration of hybrid architectures. Moreover, computational efficiency is difficult to assess in a hybrid quantum-classical hardware setting, as architecturally, the quantum GAN relies on several heavy classical computations that may remain costly under a quantum-hardware regime.

Additionally, the dataset used for the framework was limited in size due to computational resource costs, and its shapes were also segmented to the same length as the synthetic shapes for training purposes. Given the baseline coverage (COV) results, and size constraints, the dataset may limit the generalisability of the framework to real-world CAD geometries.

The models also exhibited difficulty in accurately synthesising global structures such as symmetry, likely as a result of requiring global segment coordination (which is purely an emergent property for our GAN). The architecture therefore may benefit from additional global priors or conditioning methods to better extrapolate global structures from local curvature.

## 6 CONCLUSION AND FUTURE WORK

This work introduces HMB-GAN, a fully-differentiable geometric GAN that constructs shapes through stitched multi-segment Bézier curves. Across all point cloud metrics, the quantum architecture showed modest improvements, also achieving marked reduction in model parameters and faster convergence across epochs (though at substantially higher computational cost due to simulation overhead). Although current simulated-quantum evaluation is constrained by classical hardware, these findings suggest that with improved quantum processing methods, VQCs may offer advantages in modelling nonlinear relationships in geometric domains.

## ACKNOWLEDGMENT

The authors would like to sincerely acknowledge the School of Science, Computing and Emerging Technologies, School of Engineering and School of Design and Architecture, Swinburne University of Technology for their generous support and resources throughout this research.